# Multimodal deep learning for mapping forest dominant height by fusing GEDI with earth observation data

Man Chen[a], Wenquan Dong[a,*], Hao Yu[b,*], Iain Woodhouse[a], Casey M. Ryan[a], Haoyu Liu[c], Selena Georgiou[a], Edward T.A. Mitchard[a,d]

[a] *School of GeoSciences, University of Edinburgh, Edinburgh EH8 9XP, UK*

[b] *School of Engineering, University of Edinburgh, Edinburgh EH9 3FG, UK*

[c] *School of Informatics, University of Edinburgh, Edinburgh EH8 9AB, UK*

[d] *Space Intelligence Ltd, 93 George Street, Edinburgh  EH2 3ES, UK*

**Abstract**

Accurate estimation of forest height plays a pivotal role in quantifying carbon stocks. The integration of multisource remote sensing data and deep learning models offers new possibilities for accurately mapping high spatial resolution forest height. We found that the Global Ecosystem Dynamics Investigation (GEDI) relative heights (RH) metrics exhibited strong correlation with the mean of the top 10 highest trees (dominant height) measured in situ at the corresponding footprint locations. Consequently, we propose a novel deep learning framework termed the multi-modal attention remote sensing network (MARSNet) to estimate forest dominant height by extrapolating dominant height derived from GEDI, using Setinel-1 C-band Synthetic Aperture Radar (SAR) data, Advanced Land Observing Satellite-2 (ALOS-2) Phased Array type L-band Synthetic Aperture Radar-2 (PALSAR-2) data, Sentinel-2 optical data and ancillary data. MARSNet comprises separate encoders for each remote sensing data modality to extract multi-scale features, and a shared decoder to fuse the features and estimate height. Using individual encoders for each remote sensing imagery avoids interference across modalities and extracts distinct representations. To focus on the efficacious information from each dataset, we reduced the prevalent spatial and band redundancies in each remote sensing data by incorporating the extended spatial and band reconstruction convolution (ESBConv) modules in the encoders. MARSNet achieved commendable performance in estimating dominant height, with an $R^2$ of 0.62 and RMSE of 2.82 m, outperforming the widely used random forest approach which attained an $R^2$ of 0.55 and RMSE of 3.05 m. We demonstrated the efficacy of the MARSNet modules and the expansion of data sources for improving dominant height estimation through network ablation studies and data ablation studies. Finally, we applied the trained MARSNet model to generate wall-to-wall maps at 10 m resolution for Jilin, China. Through independent validation using field measurements, MARSNet demonstrated an $R^2$ of 0.58 and RMSE of 3.76 m, compared to 0.41 and 4.37 m for the random forest baseline. Additionally, MARSNet effectively mitigated the tendency to overestimate in low height areas and underestimate in high canopy areas. Our research demonstrates the effectiveness of a multimodal deep learning approach fusing GEDI with SAR and passive optical imagery for enhancing the accuracy of high resolution dominant height estimation.

## 1. Introduction

Forests accumulate and store a substantial amount of carbon dioxide from the atmosphere, playing a crucial role in mitigating climate change (Mitchard 2018). A pivotal parameter, forest height, represents the vertical structure of forests and is directly correlated with the calculation of forest biomass, linking it to a forest's carbon storage capacity (Liu et al. 2019; Nandy et al. 2021). Precise quantification of forest height can therefore provide vital insights into a forest's carbon sequestration potential and its ability to regulate global carbon cycling (Le Toan et al. 2011). Additionally, forest height also serves as an ecological indicator of biodiversity as different species of flora and fauna inhabit different vertical layers of a forest, thus informing biodiversity assessments and conservation strategies (Oettel and Lapin 2021). Moreover, forest height has a significant impact on the local climate and hydrological cycle (He and Pomeroy 2023), and it is indispensable for evaluating forest degradation and quantifying the effectiveness of forest restoration efforts (Liang et al. 2023).

Traditional field inventory methods can provide reliable tree height information at a small scale (Andersen et al. 2006). However, this approach is labor-intensive and time-consuming (Chave et al. 2014), and it is infeasible to produce large-scale forest height products using field measurements alone. Compared to traditional methods, remote sensing technology has significantly improved the efficiency of mapping forest heights over large areas and in regions that are difficult to access (Duncanson et al. 2010; Feng et al. 2023; Lang et al. 2023; Wang et al.

2023). Satellite-based methodologies in remote sensing data for forest height estimation primarily rely on passive optical, active SAR and Light Detection and Ranging (LiDAR) data.

Optical sensors measure surface reflectance at different wavelengths related to canopy chemical and structural properties (Asner and Martin 2008). Various optical vegetation indices and spectral bands, such as Normalized Difference Vegetation Index (NDVI) or combinations of band ratios, have demonstrated strong correlations with forest height (Matese et al. 2017; Staben et al. 2018). However, the usability of optical data can also be reduced by atmospheric effects from clouds and haze (Makarau et al. 2014; Meraner et al. 2020). Additionally, optical data cannot directly provide vertical forest structure parameters, as they only capture information from the top canopy surface. SAR is particularly useful in regions frequently covered by clouds due to its ability to penetrate through clouds. (Mitchard et al. 2011; Woodhouse 2017). Additionally, different frequencies can penetrate varying depths into the canopy, and the backscatter values contain information about the vegetation's vertical structure, making it one of the primary techniques for estimating forest height (Kumar et al. 2020; Lei et al. 2021). Although passive optical and SAR instruments both collect continuous imagery, optical and SAR data have been shown to saturate in relatively shorter forests (Campbell et al. 2021; Kugler et al. 2014; Rodríguez-Veiga et al. 2019; Woodhouse et al. 2012).

LiDAR data, with its strong penetration capability, is highly sensitive to even dense forests with minimal or no saturation (Duncanson et al. 2022), making it one of the most effective remote sensing techniques for measuring the three-dimensional structure of forests (Coops et al. 2021). LiDAR actively emits laser pulses, relying on the penetration of the pulses to detect the three-dimensional structure of vegetation (Guo et al. 2020). Airborne laser scanning (ALS) can provide continuous and precise estimations of forest height at the regional scale (Mielcarek et al. 2020; Tymińska-Czabańska et al. 2022). However, ALS data are costly and often limited in coverage. The advent of space-borne LiDAR, such as GEDI, has the potential to overcome these constraints by offering large-scale, globally comprehensive data on terrain and vegetation height, thus becoming the preferred dataset for extensive mapping applications (Dubayah et al. 2020; Liu et al. 2021). GEDI is a mission aimed at creating detailed 3D mapping of the Earth's forests and topography using a LiDAR system from the International Space Station (ISS) (Dubayah et al. 2020). GEDI provides an unprecedented sampling density of forest structural properties (Potapov et al. 2021); however, its data consists of discrete footprints, limiting continuous spatial analysis at high resolution.

To overcome the spatial discontinuity problem of space-based LiDAR, such as GEDI, researchers have attempted to integrate satellite LiDAR with continuous satellite imagery to map wall-to-wall forest canopy height (Ceccherini et al. 2023; Liu et al. 2022). This is often achieved by establishing regression models between LiDAR as the reference data and continuous satellite imagery. Researchers often utilize machine learning methods model these data, as they are frequently highly non-linear and correlated with each other (Luo et al. 2020). Machine learning methods such as k-nearest neighbors, support vector machines (SVMs), and random forests have emerged as popular techniques for extrapolating LiDAR data due to their stable performance (Baccini et al. 2012; Matasci et al. 2018; Tang et al. 2018; Zhang et al. 2013). More recently, gradient boosting algorithms, such as extreme gradient boosting (XGBoost) and light gradient boosting machine (LightGBM), have also gained widespread usage, due to their rapid training speeds and interpretability features (Shendryk 2022; Song et al. 2023). However, these methods do not utilize the spatial information of remote sensing imagery. These methods rely on statistical relationships between predictor variables rather than their explicit spatial context. Particularly when integrating with 10 m Sentinel-1 and Sentinel-2 data to generate canopy height or biomass maps, using the mean values of pixels within footprints to establish relationships with LiDAR data such as from GEDI, ignores the textural information within the footprints. Unlike traditional methods like random forests that use aggregated pixel statistics, deep learning approaches are capable of exploiting the full texture and spatial patterns within satellite image patches (Reichstein et al. 2019). Lang et al. (2023) employed a convolutional neural network (CNN) to extrapolate GEDI data into 10 m canopy height map, incorporating Sentinel-2. Dong et al. (2023) employed an attention UNet architecture to integrate multi-source remote sensing data and GEDI, to produce 10 m aboveground biomass (AGB) maps. However, these studies have merely superimposed multimodal information. Due to the heterogeneous complexities inherent within various remote sensing modalities, such approaches fail to effectively leverage the rich information available from multimodal data (Ma et al. 2023; Maimaitijiang et al. 2020).

In this study, we proposed a novel multimodal attention remote sensing network (MARSNet) to fully exploit the synergies across various remote sensing modalities for extrapolating GEDI data. Our objectives are threefold: 1) to investigate the relationships between GEDI RH metrics and field-measured forest dominant height to convert GEDI RH to ground-truth dominant height, 2) to examine the performance of MARSNet in estimating forest dominant height, 3) to explore the potential of MARSNet through ablation studies.

## 2. 2. Data and Materials

### 2.1. Study Area

Jilin Province, situated in northeastern China between 121°38′E–131°19′E, and 40°52′N–46°18′N, encompasses a total land area of 187,000 km$^2$, which constitutes 2% of China's territorial extent. The province boasts varied topographical features, descending from southeast to northwest, and includes eastern mountainous terrains as well as central and western plain regions. The province exhibits a temperate continental monsoon climate, typified by prolonged frigid winters and brief hot summers. Mean annual temperature ranges from 3-5°C, while annual precipitation averages 763.3 mm (Liu et al. 2018). Significantly, Jilin Province stands out as one of the most forested areas in China, with forests spanning approximately 7.85 million hectares, accounting for 41.9% of the province's total land area, which is markedly higher than the national average forest coverage rate of 22.96% (Xu et al. 2019). Additionally, the forest areas in Jilin Province primarily consist of temperate coniferous and broad-leaved mixed forests. The study area is also rich in biodiversity, hosting 408 main tree species and more than 250 types of woody plants. Dominant tree species, including oak, coniferous, larch, and poplar, constitute over 90% of the total stand area in Jilin Province (Shi et al. 2021).

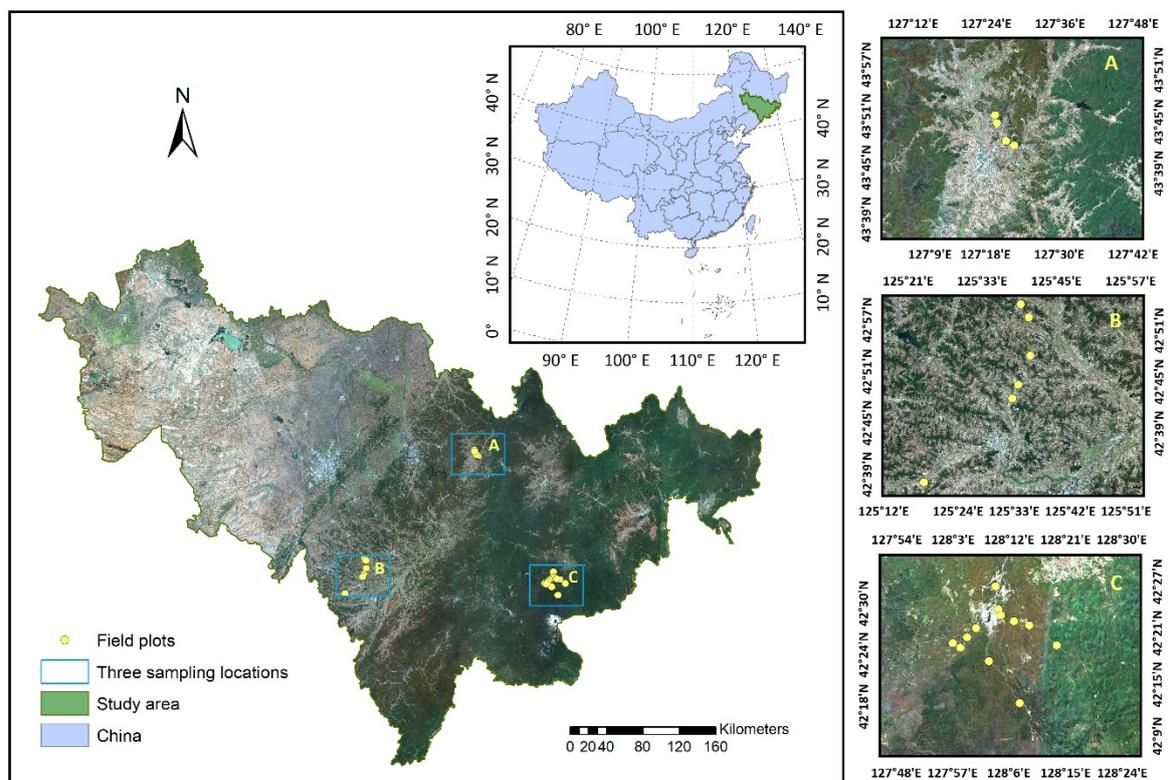

**Fig. 1.** Study area and field plot. (A-C) Enlarged maps of the three sampling locations. True color Sentinel-2 images of the overall study area and the three specific sampling locations.

### 2.2. Field Data

The field data collection was conducted in Jiaohe (A), Dongfeng (B), and ChangbaiShan (C), as displayed in Fig. 1. These locations were strategically chosen to ensure a more uniform and representative spatial distribution of samples across the Jilin province. Tree heights were measured across 24 plots using the Vertex IV and Transponder T3 instruments, of which 22 were under GEDI Version 1 footprints. Each plot is a circle with a diameter of 25 m, consistent in size and shape with the GEDI footprint. This allows the ground truth data to be directly comparable to GEDI data, without the need for additional scaling or interpolation. Among these 22 plots under GEDI footprints, the average number of trees with a diameter at breast height (DBH) exceeding 5 cm was 49 per plot, with a minimum of 20 trees and a maximum of 91 trees. Furthermore, to circumvent spatial autocorrelation and guarantee the independence of samples, each sampling plot was strategically positioned at a minimum distance of 2 km from any other plot. In this study, we computed the mean height of the tallest 10 trees within each plot as the dominant height for further analysis with remote sensing data. The dominant height holds greater significance in these heterogeneous forests, as it takes into account the largest trees. In mixed or uneven-aged forests, the height of dominant individuals is more indicative of site productivity (Morin et al. 2022).

*2.3. Remote sensing data*

*2.3.1. GEDI*

GEDI is equipped with three LiDAR lasers, one of which is split into dual beams known as "cover beams," while the other two lasers operate as "full power beams." The distinction between the two kinds of beams lies in their penetration capabilities: cover beams and full power beams can penetrate approximately 95% and 98% of the forest canopy to reach the ground, respectively (Spracklen and Spracklen 2021). GEDI provides high-quality footprint-size waveforms, with footprints separated by approximately 60 m along-track and around 600 m between eight orbits. Each footprint covers an area with a diameter of 25 m, effectively providing detailed waveforms of forest attributes.

In this study, we utilized the GEDI Version 2 Level-2A (L2A) product, which provides vertical structure profiles for each footprint. Specifically, it incorporates RH metrics, which represent canopy height percentiles and are essential for assessing forest structure and height (Duncanson et al. 2022). We preprocessed and downloaded the GEDI data acquired in 2021 for Jilin Province on the Google Earth Engine (GEE), a cloud-based platform optimized for geospatial analysis and large-scale data processing. This preprocessing primarily involved eliminating low-quality, degraded, and daytime-collected (Beck et al. 2021) footprints through the layers provided in the L2A dataset. In order to estimate forest height more accurately, we only used GEDI's power beams (Duncanson et al. 2020; Lahssini et al. 2022; Liu et al. 2021). To further select high-quality GEDI footprints, we also filtered the footprints by combining GEDI L2B and Sentinel-2 data with the following filter conditions:
1) For GEDI footprints with a canopy cover < 0.8, those with a sensitivity of ≥ 0.95 were retained for further analysis. Conversely, for footprints with a canopy cover ≥ 0.8, a more stringent sensitivity threshold of 0.98 or higher was applied for retention. The sensitivity of a GEDI footprint refers to the maximum canopy cover that can be penetrated to detect the ground (Hancock et al. 2019). This differential sensitivity threshold, while retaining as many footprints as possible, also improves the overall reliability.
2) Footprints where the absolute difference between the GEDI L2B canopy cover and the NDVI derived from Sentinel-2 exceeded one standard deviation from the mean were removed. This is because a large absolute difference may be caused by geolocation errors.

*2.3.2. Sentinel-2*

Sentinel-2 is an Earth observation mission launched by the European Space Agency (ESA) as part of Copernicus Programme, provides high-resolution optical imagery. Its unique spectral capabilities, with 13 spectral bands, span from the visible and near-infrared (VNIR) to the shortwave infrared (SWIR) portions of the spectrum, offering comprehensive spectral sampling. We utilized Sentinel-2 Level-2A imagery from 2021, specifically selecting images acquired between April and October. This period was selected due to the presence of snow cover in Jilin Province during other months, which could obscure ground features and distort vegetation indices. Cloud masking was applied to each image using the 'QA60' band to identify and remove pixels affected by cloud or cirrus. The Normalized Difference Vegetation Index (NDVI) and Enhanced Vegetation Index (EVI) were derived because these vegetation indices are sensitive to the relative abundance and spatial distribution of vegetation (de la Iglesia Martinez and Labib 2023). The composite image was generated using the median values of the spectral bands and vegetation indices, providing a robust and representative dataset by minimizing temporal variations. In addition to the median, we also extracted the maximum, minimum, and the difference between these values of NDVI over the image collection, capturing seasonal vegetation dynamics.

*2.3.3. Sentinel-1*

Sentinel-1, a component of the Copernicus Programme, comprises European radar imaging satellites launched in 2014 and 2016, equipped with a C-band (5.405 GHz) SAR instrument. In this study, we utilized data from Sentinel-1 collected between April and October of 2021. We specifically chose this timeframe because the backscatter from SAR in other months might be adversely affected by frozen conditions, such as freezing trees and snow cover, leading to potentially lower and less reliable readings (Santoro et al. 2009). The Sentinel-1 SAR GRD data utilized in this study is a radiometrically calibrated and orthorectified product. To reduce speckle noise in the imagery, we applied a filter using a circle with a radius of 50m. Due to the incomplete provincial coverage of Jilin by data acquired from ascending satellite orbits, we utilized only SAR backscatter measurements from the descending orbits in this study. Specifically, the 10th, 50th and 90th percentile values of the VV and VH polarizations and their ratios were derived and adopted as input layers.

*2.3.4. ALOS-2 PALSAR-2*

ALOS-2 PALSAR-2 is a Japanese spaceborne SAR system, operates at L-band frequencies, ensuring deeper penetration into the forest canopy than C-band. Consequently, it yields valuable insights into forest structure and height. In this study, we used open access PALSAR-2 yearly mosaic data for 2021, providing 25m annual, consistent, and wide-area observations. HV and HH polarization modes and their ratio were used to capture the scattering characteristics of the forest. The local incidence angle, which directly influences the strength of the radar return signal, was also extracted. The backscatter values in the ALOS-2 PALSAR-2 data were converted to gamma naught (decibels value) using the following function:

$$\gamma^0 = 10 \log_{10}(DN^2) - 83 dB \tag{1}$$

where DN is the backscatter value. This conversion facilitates the comparison of backscatter values over large areas and time periods. This dataset was resampled to a 10m resolution using bicubic interpolation to match the rest of the remote sensing data.

*2.3.5. forest cover*

The European Space Agency WorldCover 10m 2021 product provides global 10m land cover classification based on Sentinel-1 radar and Sentinel-2 optical data (Zanaga et al. 2022). We acquired and processed the WorldCover dataset on Google Earth Engine (GEE) to extract forest cover information for Jilin Province, China. By clipping the data to the study area boundary and reclassifying values, we generated a 10m resolution binary forest/non-forest mask. By utilizing the binary forest/non-forest mask, we extracted GEDI footprints located within forested areas, which proved instrumental for our subsequent experimental research. Additionally, by applying the forest mask to the forest height maps, we were able to focus our analysis specifically on forested regions and generate precise maps of estimated forest height.

*2.3.6. Ancillary datasets*

In this study, we utilized NASA's High-Resolution Digital Elevation Model (NASADEM) to obtain precise topographic representation of the study area in Jilin Province, China. NASADEM is an improved digital elevation model developed by reprocessing Shuttle Radar Topography Mission (SRTM) data using advanced algorithms and additional observations from ICESat GLAS and ASTER instruments (NASA 2020). Significant enhancements include reduced data voids through improved phase unwrapping, the utilization of ICESat GLAS data for calibration, the mitigation of height ripples from the original data collection, and the seamless integration of ASTER-derived GDEM to fill remaining voids, culminating in comprehensive DEM products. Beyond elevation, we derived slope data from NASADEM to characterize topographic variations across the landscape. Slope is a pivotal parameter in forest height estimation, as it can influence the reflectance of LiDAR and radar signals. By incorporating slope, we could better account for topography's effect on remote sensing measurements. In addition, gridded longitude and latitude data were incorporated as input layers to provide explicit spatial information to the models. To be consistent with other datasets, the ancillary datasets were also resampled to 10m resolution for further analysis.

*2.4. MARSNet*

In this study, a multimodal network, termed the multi-modal attention remote sensing network (MARSNet), is proposed, as illustrated in Fig. 2. The network utilizes remote sensing datasets from Sentinel-1, Sentinel-2, ALOS-2 PALSAR-2 and ancillary data, which have a total of 34 unique bands, to estimate dominant tree heights. The task of estimating dominant height is formulated as a regression problem, with the mathematical expression provided as follows:

$$f^*(\theta) = \arg\min_{\theta} \frac{1}{K} \sum_{i=1}^{K} \|f(X_i | \theta) - Y_i\|_2^2 \tag{2}$$

where $f^*()$ represents the MARSNet, with $f(X_i|\theta)$ indicating the estimated dominant height and $Y_i$ denoting the dominant height derived from GEDI. $K$ represents the total number of samples within the dataset. $\theta$ represents the model parameters that are updated during the backpropagation process.

MARSNet is based on an encoder-decoder architecture, comprising four encoders with associated modal fusion modules, and one decoder. Specifically, remote sensing modalities are input into their corresponding encoders, each of which generates a set of feature maps with dimensions changing as follows — 64 × 64 × 64, 128 × 32 × 32, 256 × 16 × 16, 512 × 8 × 8 (band × height × width). The utilization of a shared encoder may result in inter-module interference. This interference arises because the trained parameters of the encoder are shared during model training, potentially compromising its capacity to effectively extract features from each modality. Consequently, in MARSNet, the employment of four separate, identical encoders to extract multi-scale representative features from each modality independently is proposed. Subsequent operations employ the modal fusion module to fuse the feature maps with the same dimension from different modalities, which are then utilized in the decoder to estimate dominant height.

Detailed descriptions of the modules are below.

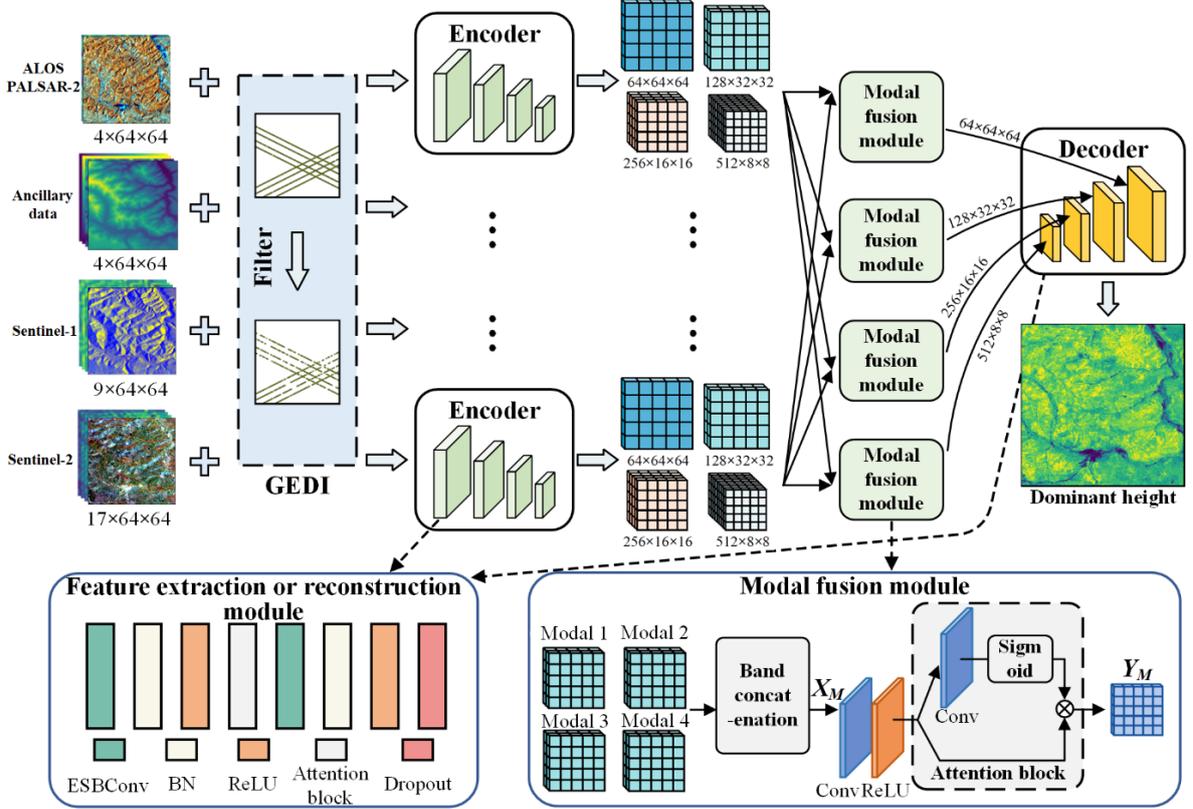

**Fig. 2.** The architecture of the proposed multi-modal attention remote sensing network (MARSNet).

### 2.4.1. Encoder and Decoder

As depicted in Fig. 2, the encoder and the decoder have the mirroring architecture, responsible for latent feature extraction and height reconstruction respectively. The encoder/decoder network involves multiple sub-blocks, with each encompassing an extended spatial and band reconstruction convolution (ESBConv) module, a batch normalization (BN) layer, a rectified linear unit (ReLU), and an attention layer. To thwart overfitting and amplify the generalization capabilities of the model with respect to unobserved remote sensing data, a 0.25 dropout layer is incorporated into the final layer. The encoder efficiently extracts multi-scale features from various modalities of remote sensing data, whereas the decoder effectively reconstructs the fused, multi-scale modal information, ultimately estimating the dominant height. This design ensures the efficacious fusion of various types of remote sensing data, facilitating robust multi-modal feature extraction and reliable reconstruction.

### 2.4.2. ESBConv

We extended the original spatial and channel reconstruction convolution (SCConv) operator (Li et al. 2023), which is a plug-in convolution module. The module, named the extended spatial and band reconstruction convolution module, is designed to augment the performance of convolutional neural network (CNN) models by reducing the prevalent spatial and band redundancies in features. As depicted in Fig. 3, ESBConv is composed of three distinct components: a standard 1 × 1 convolution for band adjustment, a spatial reconstruction unit (SRU), and a band reconstruction unit (BRU). These components are arranged in sequence. The input features, $X$, first pass through a 1 × 1 convolution layer for band adjustment. They then proceed to the SRU, yielding spatial-refined features, $X^w$. Finally, after processing by the BRU, band-refined features, $Y$, are produced. Remote sensing data from each modality is voluminous and complex. The significance of each modality may vary when processed by different kernels. By employing the SRU and BRU, one can effectively eliminate spatial and band redundancy in the data, extracting more representative modality features, which in turn enhances the dominant height prediction performance of the model.

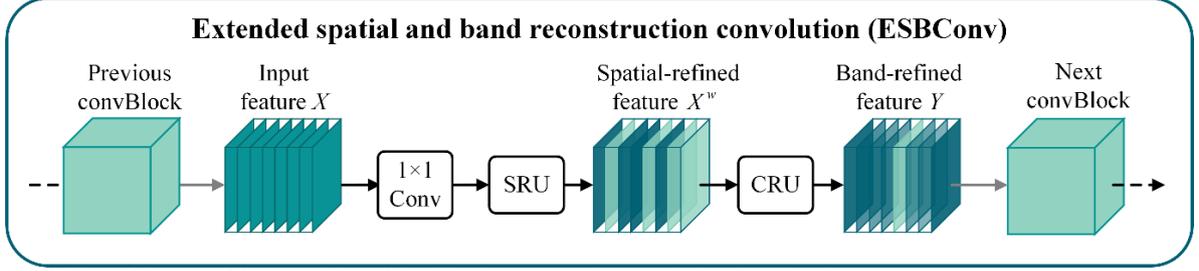

**Fig. 3.** The architecture of the extended spatial and band reconstruction convolution (ESBConv).

*2.4.2.1. SRU for spatial redundancy*
The SRU employs a *Separate-Reconstruct* operation, as illustrated in Fig. 4.

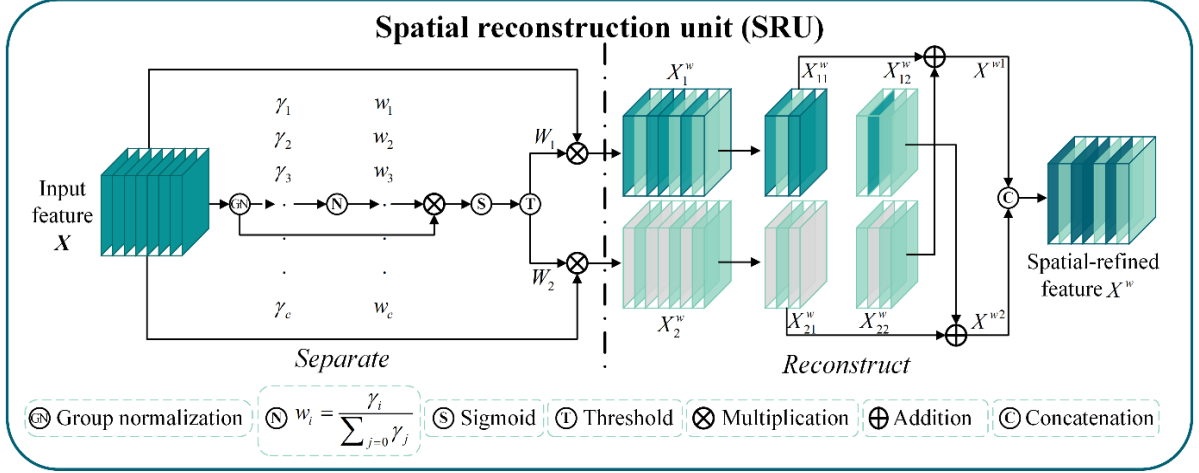

**Fig. 4.** The structure of the spatial reconstruction unit (SRU).

**Separate:** The objective of the separation operation is to segregate feature maps with considerable informative content from those with limited informative content, with respect to the spatial context. The input feature maps of the SRU, denoted as $X$, are standardized by the group normalization (GN) layer and subsequently operated upon by the normalized correlation weights $W_\gamma \in R^b$. The entire procedure for acquiring $W$ can be formulated as follows:

$$W = Gate\left(Sigmoid\left(W_\gamma(GN(X))\right)\right) \quad (3)$$

Herein, the *Sigmoid* function serves as the activation function and the *Gate* function acts as a threshold. Weights $W_1$ are derived by assigning a value of 1 to elements above the threshold, while a value of 0 is assigned to derive the non-informative weights $W_2$. Finally, the two weighted features, namely the informative $X_1^w$ and less informative $X_2^w$, are obtained by multiplying $X$ by $W_1$ and $W_2$, respectively.

$$\begin{aligned} X_1^w &= W_1 \otimes X \\ X_2^w &= W_2 \otimes X \end{aligned} \quad (4)$$

where $\otimes$ is the Hadamard product.

**Reconstruct:** The reconstruction operation combines features with abundant information and those with limited information to produce more informative features while conserving space. Specifically, a cross-reconstruction approach is utilized, wherein two distinct information features are merged after weight application, yielding $X^{w1}$ and $X^{w2}$. Subsequently, these are then connected to obtain the spatially refined feature maps $X^w$. The entire process can be formulated as follows:

$$\begin{aligned} X_{11}^w \oplus X_{22}^w &= X^{w1} \\ X_{21}^w \oplus X_{12}^w &= X^{w2} \\ X^{w1} \cup X^{w2} &= X^w \end{aligned} \quad (5)$$

where $\oplus$ is the element-wise addition, and $\cup$ is the concatenation operation.

*2.4.2.2. BRU for band redundancy*

The BRU, which employs a *Split-Transformer-Fuse* architecture as illustrated in Fig. 5, is utilized to reduce redundancy in spatial-refined feature maps along the band dimension. Furthermore, it uses lightweight convolutional operations to extract rich representative features from the input, concurrently processing redundant features with efficient operations and feature reuse schemes.

**Split:** Given $X^w \in \mathbb{R}^{b \times h \times w}$, the spatially refined features obtained from the SRU, we derive $\alpha B$ bands and $(1-\alpha)B$ bands from $X^w$ through a splitting operation. Here, $\alpha$ represents a split ratio constrained to $0 \leq \alpha \leq 1$. Subsequently, the bands of feature maps are compressed utilizing a $1 \times 1$ convolution kernel, yielding $X_{up}$ and $X_{low}$, respectively.

**Transformer:** $X_{up}$ are fed into the upper transformation branch, functioning as the "Rich Feature Extractor" (RFE), and are then subjected to the group-wise convolution (GWC) and point-wise convolution (PWC), respectively. The merits of GWC and PWC are evidenced in (Krizhevsky et al. 2012; Kundu et al. 2019). The results from these convolutions are summed to obtain the output $Y_1$. The corresponding mathematical expression is as follows:

$$Y_1 = M^G X_{up} + M^{P_1} X_{up} \qquad (6)$$

Let $M_{\frac{\alpha b}{b}}^G \in \mathbb{R}^{\frac{\alpha b}{gr} \times k \times k \times b}$ and $M^{P_1} \in \mathbb{R}^{\frac{\alpha b}{r} \times 1 \times 1 \times b}$ denote the learnable weight matrices for GWC and PWC. $X_{up} \in \mathbb{R}^{\frac{\alpha b}{r} \times h \times w}$ and $Y_1 \in \mathbb{R}^{b \times h \times w}$ represent the input and output of the upper branch, respectively.

In the lower transformation branch, $X_{low}$ is utilized in a twofold manner: initially to generate shallow feature maps via $1 \times 1$ PWC operations as a complement to RFE, and simultaneously $X_{low}$ is refused to extract more feature maps without incurring extra computational costs. These generated and reused features are then concatenated, forming the output $Y_2$ for the lower stage, as delineated below:

$$Y_2 = M^{P_2} X_{low} \cup X_{low} \qquad (7)$$

where $M^{P_2} \in \mathbb{R}^{\frac{(1-\alpha)b}{r} \times 1 \times 1 \times (1-\frac{(1-\alpha)}{r})b}$ represents a learnable weight matrix of PWC. $X_{low} \in \mathbb{R}^{\frac{(1-\alpha)b}{r} \times h \times w}$ and $Y_2 \in \mathbb{R}^{b \times h \times w}$ are the input and output feature maps of the lower branch, respectively.

**Fuse:** The Selective kernel networks (SKNet) (Li et al. 2019) method is employed for the adaptive fusion of the output features $Y_1$ and $Y_2$. Initially, global average pooling is utilized to combine global spatial information with band-wise statistical information, resulting in global band-wise descriptors $S_1$ and $S_2$. Subsequently, band-wise soft attention operations are applied to the stacked $S_1$ and $S_2$, yielding the weight vectors $\beta_1$ and $\beta_2$. Finally, the band-refined features $Y$ are computed using the feature weight vectors, as formulated below:

$$Y = \beta_1 Y_1 + \beta_2 Y_2 \qquad (8)$$

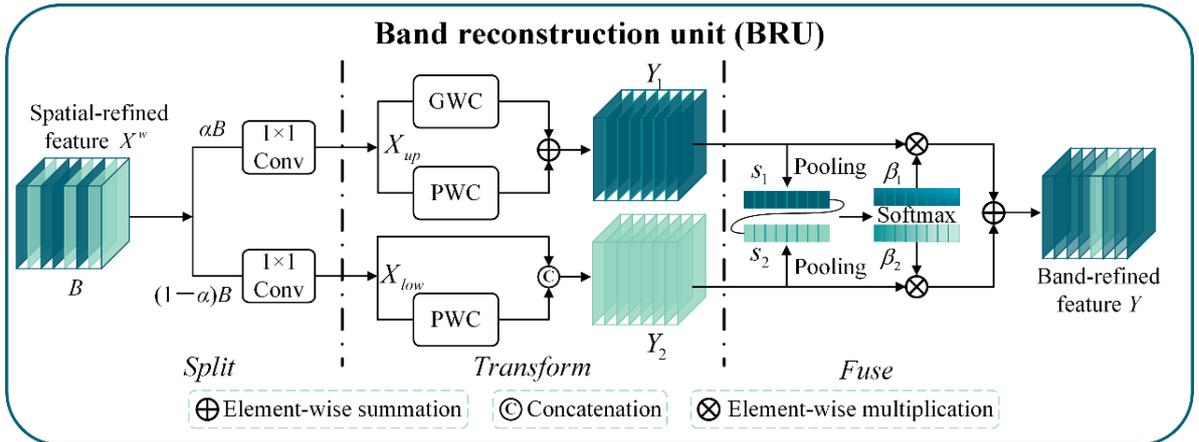

**Fig. 5.** The structure of the band reconstruction unit (BRU).

### 2.4.3. Model fusion module

As depicted in Fig. 2, the modal fusion module employs a two-step fusion strategy for different modalities obtained from the encoder: initial concatenation along the band dimension, followed by a fusion through the spatial attention operator. Firstly, the band concatenated feature maps, denoted as $X_M$, undergo a sequence of operations — passing through a $3 \times 3$ convolutional layer, undergoing ReLU activation, proceeding through another $3 \times 3$ convolutional layer, and finally, applying the *Sigmoid* activation — to generate attention weight maps. These weight maps are subsequently utilized to spatially weigh the ReLU output feature map, thereby emphasizing the critical modal regions identified by the model.

*2.4.4. Model Training and Implementation*

*2.4.4.1. Loss Function Definition*
We formulated the task of predicting dominant height as a regression problem. The loss function used in our model is defined as:

$$\mathcal{L}_{loss} = \mathcal{L}_{MSE} + \lambda \|\theta\|^2 \tag{9}$$

where the first term is the mean squared error (MSE) loss and the second term is an $l_2$ regularization with regularization parameter λ.

*2.4.4.2. Data Preprocessing*
The GEDI data, captured in 25-meter diameter circles, poses a challenge for use as labels in deep learning due to misalignment with remote sensing image pixels. To integrate GEDI and image data, we adopted an intersection-based approach. Specifically, if the centroid of a pixel lies within a GEDI footprint, that pixel is assigned the corresponding tree height value. This mitigates potential boundary mismatches between the GEDI and image data, ensuring more cohesive alignment for analysis. Subsequently, both the labels and the original remote sensing images were segmented into patches of size 64×64 pixels. These patches were randomly partitioned into training, validation, and test sets at a ratio of 8:1:1. As a result, the training set contains 85,150 samples, while the validation and test sets have 10,645 and 10,643 samples respectively. Z-score standardization (Yu et al. 2022) was applied to the data to ensure a consistent scale, thereby enhancing the robustness and convergence properties of the subsequent training process.

*2.4.4.3. Implementation Details*
The deep learning models were implemented on a dedicated server equipped with an NVIDIA RTX A5000 24GB GPU. The initial learning rate was set to 0.001, with training constrained to a maximum of 50 epochs. The early stopping strategy was adopted in the paper. Furthermore, a batch size of 64 was used along with an $l_2$ regularization coefficient of $1 \times 10^{-5}$ to mitigate overfitting. The Adam optimizer was used for optimization. The epoch that achieved the best performance on the validation set was selected as the final trained model for evaluation (Yu et al. 2023).

*2.5 Forest height accuracy assessment*

To facilitate a comparison between the results of MARSNet and random forest, we evaluated the accuracy of MARSNet at the footprint level, consistent with the approach used for random forest. We assessed the accuracy of forest height estimation by examining the difference between derived dominant forest height dataset and model predictions. Specific metrics utilized to quantify the error between estimated and observed forest heights included the coefficient of determination ($R^2$), root mean square error (RMSE), relative RMSE (rRMSE). These metrics are formulated as:

$$R^2 = 1 - \frac{\sum_{i=1}^{n}(x_i - y_i)^2}{\sum_{i=1}^{n}(x_i - \bar{y})^2} \tag{10}$$

$$RMSE = \sqrt{\frac{\sum_{i=1}^{n}(x_i - y_i)^2}{n}} \tag{11}$$

$$rRMSE = \frac{RMSE}{\bar{y}} * 100\% \tag{12}$$

where, $x_i$ represents the dominant forest height derived from GEDI RH98 or field measurements for the observation; $y_i$ is the predicted forest height using different algorithms models; $n$ is the number of labels in the test set; and $\bar{y}$ is the average of predicted forest height.

## 3. Result

### 3.1. Forest height retrieval form GEDI as labels

GEDI offers valuable vertical profiles of forest structure, a prevalent practice in many studies is to directly adopt RH metrics, like RH98, to produce canopy height maps (Lahssini et al. 2022; Lang et al. 2022). This widely used method may not always reflect the true on-the-ground conditions. We analysed the relationship between GEDI metrics and field-measured tree heights in order to calibrate the spaceborne LiDAR observations to actual ground conditions, enabling the generation of wall-to-wall maps depicting realistic estimates of tree height. By analysing the RH values at intervals of 5 from RH60 to RH95, as well as RH98, and comparing them with field data, we found that the relationship with the field data strengthens as the RH value increases (Table 1). GEDI RH98 exhibited the strongest relationship with all field height metrics, with largest $R^2$ and smallest rRMSE. For various field height metrics, the mean height of the top 10 trees and 15 trees per plot exhibited the strongest correlation with the GEDI RH metrics.

**Table 1**
Comparison of GEDI RH metrics with field-measured tree heights in Jilin.

|  | Largest tree height | | Mean of top 5 trees | | Mean of top 10 trees | | Mean of top 15 trees | | Mean of top 20 trees | | Mean height of all trees | |
|---|---|---|---|---|---|---|---|---|---|---|---|---|
|  | $R^2$ | rRMSE (%) | $R^2$ | rRMSE (%) | $R^2$ | rRMSE (%) | $R^2$ | rRMSE (%) | $R^2$ | rRMSE (%) | $R^2$ | rRMSE (%) |
| **RH98** | 0.63 | 17.80 | 0.73 | 14.79 | **0.76** | **14.01** | 0.76 | 14.53 | 0.74 | 15.84 | 0.47 | 22.40 |
| **RH95** | 0.59 | 18.83 | 0.68 | 15.95 | 0.72 | 15.03 | 0.73 | 15.34 | 0.72 | 16.41 | 0.47 | 22.49 |
| **RH90** | 0.46 | 21.60 | 0.51 | 19.73 | 0.54 | 19.34 | 0.55 | 20.05 | 0.53 | 21.19 | 0.31 | 25.57 |
| **RH85** | 0.38 | 23.20 | 0.42 | 21.50 | 0.48 | 20.58 | 0.50 | 20.96 | 0.50 | 21.75 | 0.33 | 25.29 |
| **RH80** | 0.29 | 24.74 | 0.32 | 23.26 | 0.40 | 22.18 | 0.43 | 22.43 | 0.44 | 22.99 | 0.33 | 25.24 |
| **RH75** | 0.24 | 25.61 | 0.25 | 24.41 | 0.32 | 23.50 | 0.36 | 23.83 | 0.38 | 24.29 | 0.33 | 25.27 |
| **RH70** | 0.21 | 26.19 | 0.21 | 25.08 | 0.27 | 24.32 | 0.31 | 24.70 | 0.34 | 25.12 | 0.31 | 25.58 |
| **RH65** | 0.15 | 27.14 | 0.15 | 26.06 | 0.20 | 25.53 | 0.23 | 26.03 | 0.26 | 26.42 | 0.26 | 26.54 |
| **RH60** | 0.10 | 27.85 | 0.11 | 26.71 | 0.15 | 26.35 | 0.17 | 26.35 | 0.20 | 27.53 | 0.20 | 27.56 |

Specifically, RH98 showed strongest correlation ($R^2$= 0.76) with mean height of the top 10 and 15 trees per plot (Table 1). Since the mean height of the top 10 trees yielded slightly lower rRMSE, its linkage with RH98 is optimal (Fig. 6). This indicates that in this region, the mean height of the top 10 trees (dominant height) could represent the dominant height of the upper canopy fraction within a 25m diameter circle most closely aligned with the relative canopy heights derived from the GEDI LiDAR waveforms. The GEDI RH98 can be converted into dominant height using the following function:

$$y_d = 0.73 x_{RH98} + 7.86 \qquad (13)$$

where $x_{RH98}$ represents different GEDI RH98, $y_d$ represents dominant height.

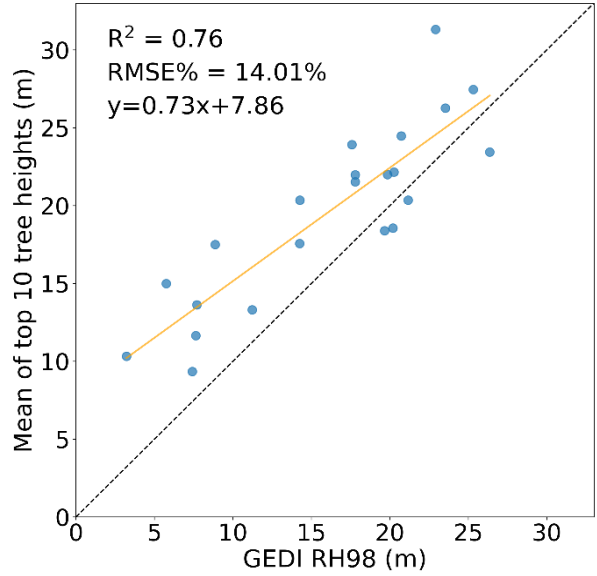

**Fig. 6.** Scatter plot between GEDI RH 98 and mean of top 10 tree heights in Jilin.

*3.2. Comparison between MARSNet and random forest*

Dominant height derived from GEDI LiDAR data was utilized to estimate forest dominant height across Jilin province, China. In order to evaluate the performance of the proposed MARSNet for dominant height estimation, comparative experiments were conducted against the widely used RF algorithm. The accuracy of MARSNet is notably superior to that of the RF model. Both models tend to overestimate when the dominant height is relatively low and underestimate when the dominant height is higher. However, this trend is less pronounced in the MARSNet model compared to the Random Forest model. It could be seen in Fig. 7, the fit of our MARSNet model aligns more closely with the 1:1 line compared to the random forest model. he MARSNet estimates exhibit a tighter clustering around the ideal linear relationship, while the random forest exhibits a higher deviation from the 1:1 line and more dispersed estimates. Quantitatively, the RF model obtained an $R^2$ of 0.55 and RMSE of 3.05m on the test data. In comparison, MARSNet attained an improved $R^2$ of 0.62 and lower RMSE of 2.82m.

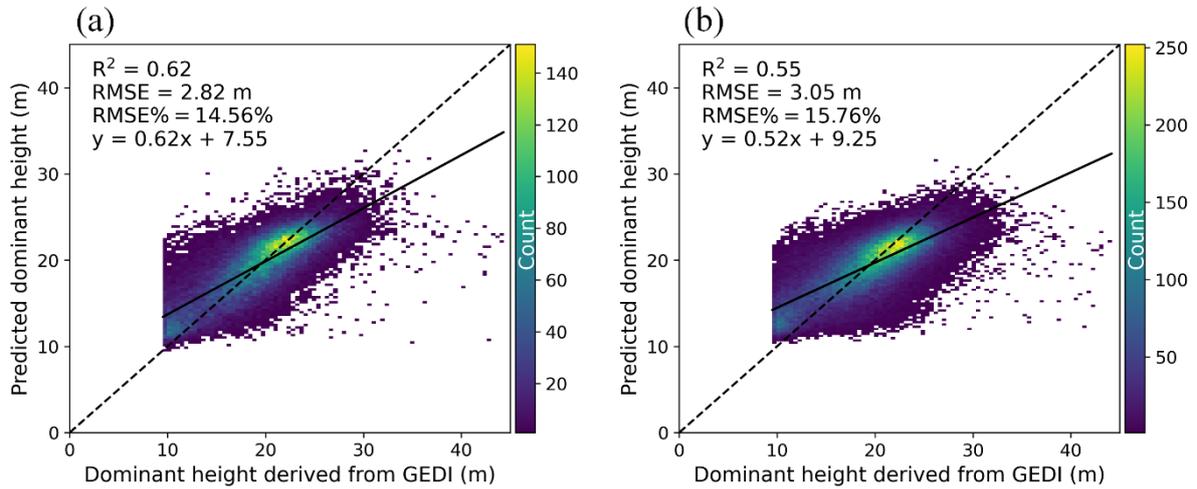

**Fig. 7.** Comparison of the performance of (a) MARSNet and (b) Random forest.

*3.3. Ablation Study of MARSNet*

*3.3.1. Ablation study of MARSNet modules*

To validate the performance gains achieved by the proposed MARSNet architecture over the random forest model, ablation studies were conducted to analyse the impact of different network components. When using a shared encoder for all four remote sensing modalities without the ESBC module, the model achieved an $R^2$ of 0.57 and RMSE of 2.99 m (Fig.8 a). This performance is slightly better than that of the Random Forest but is notably inferior to MARSNet. When each of the four remote sensing data is assigned its own encoder without the use of the ESBC, the $R^2$ improves to 0.60 and the RMSE reduces to 2.89 m (Fig.8 b). Keeping a shared encoder but incorporating the ESBC module attained an $R^2$ of 0.59 and RMSE of 2.89 m (Fig.8 c). Additionally, we also experimented with using three encoders, combining the two SAR data types, Sentinel-1 and PALSAR-2, to share one encoder. The results yielded an $R^2$ of 0.61 and an RMSE of 2.83 m, a performance closely mirroring that of our MARSNet (Fig.8 d). These results demonstrate that increasing model complexity within the proposed MARSNet framework enhances estimation accuracy, and additionally, the fit of the resulting model aligns more closely with the 1:1 line. This also validates the efficacy of the individual components used for our MARSNet.

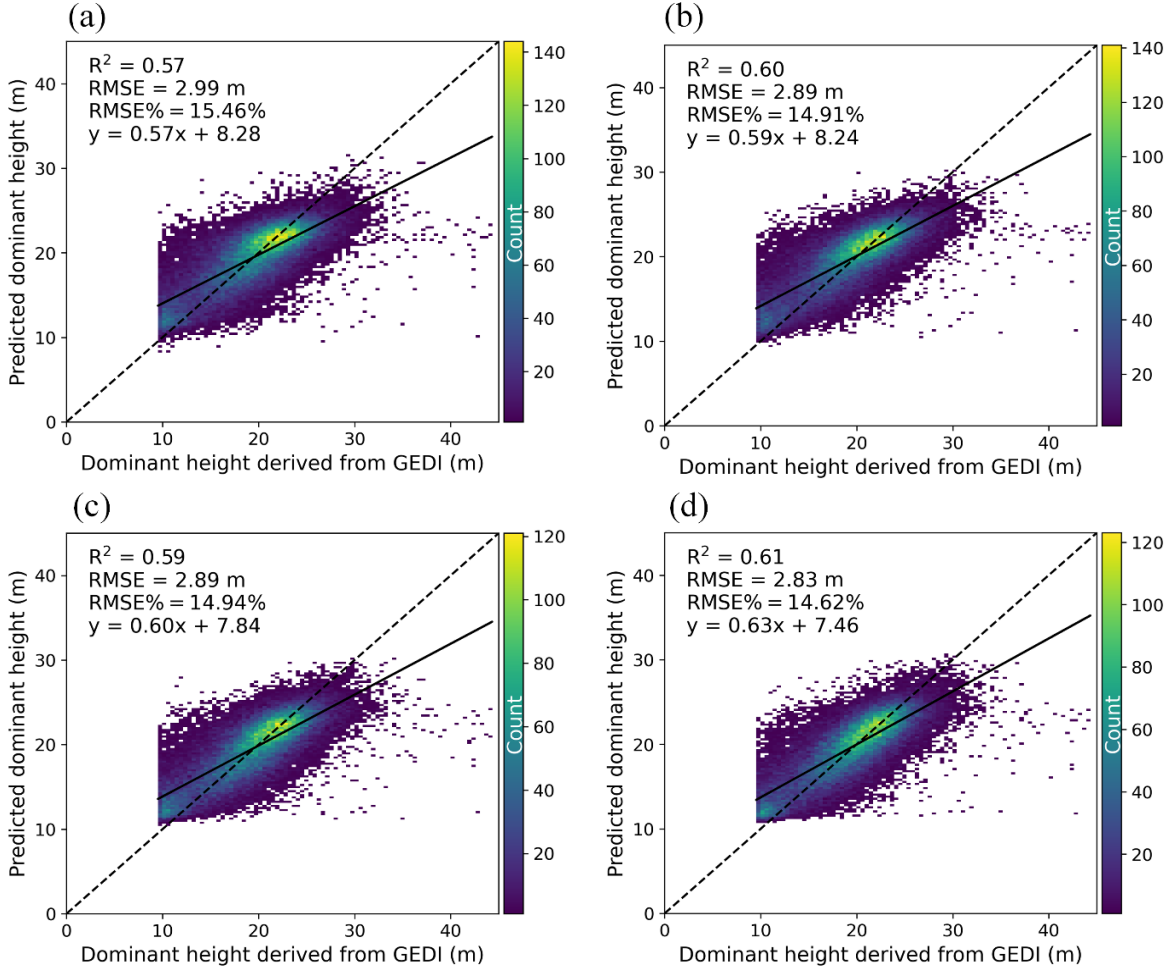

**Fig. 8.** Scatter plots for ablation study of MARSNet modules. (a) One encoder without ESBC, (b) Four encoders without ESBC, (c) One encoder with ESBC, (d) Three encoders with ESBC.

*3.3.2. Ablation study of remote sensing data modalities*

Additionally, to further assess the performance of our proposed MARSNet, we conducted experiments by incrementally removing the input remote sensing data modalities and then evaluated the resultant accuracy. Due to the large number of possible combinations from removing subsets of modalities, not every permutation was tested. Instead, we experimented with using one modality, two modalities, and three modalities, with each modality having its own dedicated encoder, to contrast with our proposed four-modal MARSNet. As expected, the accuracy was lowest when only one modality of remote sensing data (Sentinel-2) was used, with an $R^2$ of 0.56 and an RMSE of 3.02 m (Fig. 9 a). Notably, this still slightly outperformed the random forest baseline using all the four remote sensing data. When both Sentinel-1 and Sentinel-2 data were utilized, there was an improvement in accuracy, with an $R^2$ of 0.58 and an RMSE of 2.95 m (Fig. 9 b). By incorporating three types of remote sensing data, specifically Sentinel-1, Sentinel-2, and PALSAR-2, the accuracy further improved, achieving an $R^2$ of 0.60 and an RMSE of 2.89 m (Fig. 9 c).

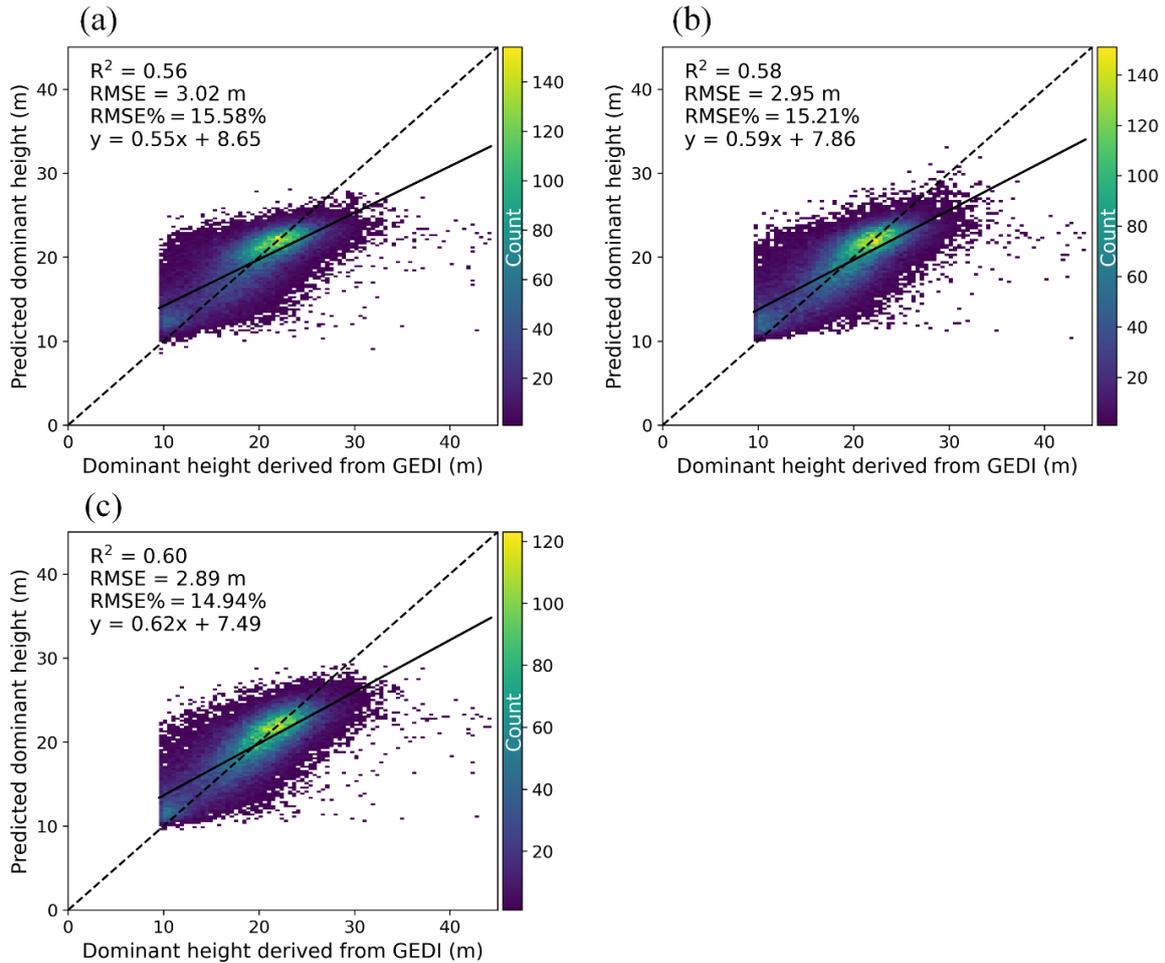

**Fig. 9.** Scatter plots for ablation study of remote sensing data modalities. (a) One modality, (b) Two modalities, (c) Three modalities.

*3.4. Independent accuracy assessment*

    The accuracy of the wall-to-wall forest dominant height maps generated by MARSNet and random forest was validated against 24 independent field measurements. As illustrated in Fig. 10, the dominant height predicted by MARSNet exhibits a strong correlation with the field-measured dominant height, achieving an R² of 0.58 and an RMSE of 3.76 m. In contrast, the dominant height predicted by the Random Forest model, when compared with the field measurements, yields an R² of 0.41 and an RMSE of 4.37 m. The linear regression fit between MARSNet predicted and field-measured dominant height had a steeper slope of 0.55, indicating its predictions more closely tracked changes in observed height values across the data range. Meanwhile, the flatter slope of 0.36 shows random forest predictions were less sensitive to changes in measured heights. The positive intercepts indicate that both models tend to overestimate the actual values in the lower range. However, the lower intercept of MARSNet suggests that its predictions were generally less biased overall compared to the random forest model.

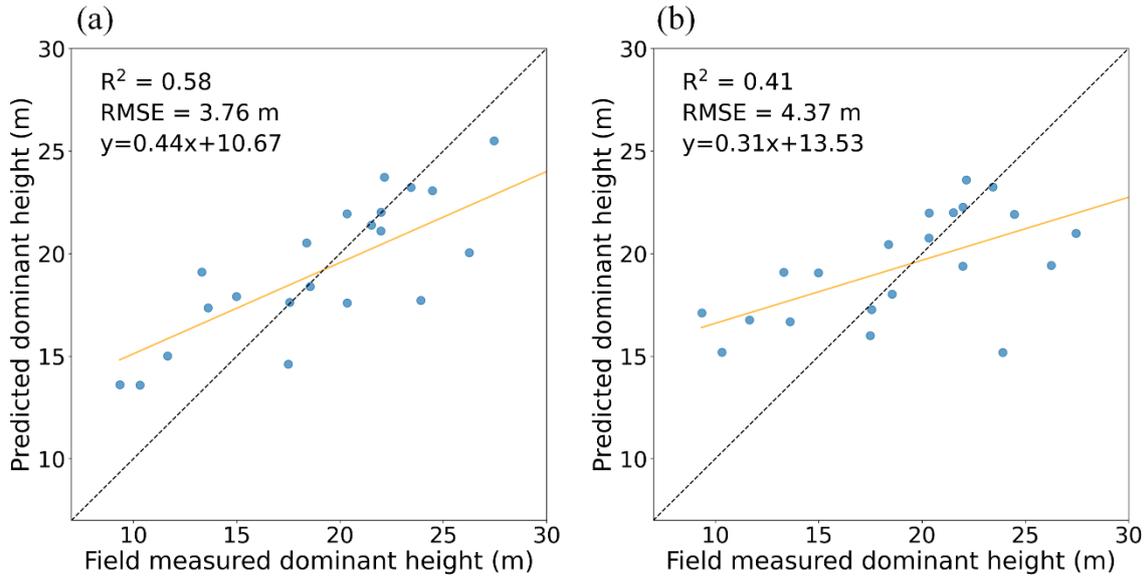

**Fig. 10.** Accuracy assessment of modeled dominant height using field measurements. (a) MARSNet, (b) Random forest.

*3.5. Spatial distribution of forest dominant height*

The wall-to-wall forest dominant height maps of Jilin province at 10 m resolution produced by the proposed MARSNet and traditional random forest model was illustrated in Fig. 11. The forests in Jilin Province are primarily concentrated in the eastern part, in the higher altitude mountainous regions (Fig. 11 (d)). Overall, the forest dominant height maps produced by both models exhibit a high degree of consistency. The maps from both models showed taller trees primarily distributed in the central part of the eastern region. Differences between the two models were visualized in Fig.11 (c), which showed a map of deviations obtained by subtracting the random forest predictions from the MARSNet results. It can be observed that in most areas the predicted dominant heights were similar between the two models.

The estimated mean forest dominant heights for Jilin province are quite similar between the two models, with MARSNet estimating an average of 20.16 m and random forest estimating 20.01 m. However, frequency distribution histograms reveal distinct differences in the distribution of estimated heights between the models (Fig. 12). The random forest model shows a higher concentration of estimates around the mean height, particularly around 20 m. In contrast, MARSNet displays markedly higher frequencies at the extremes - both in the lower height bin below 15 m and in the taller forest bin above 25 m. This distribution is also more closely aligned with the dominant height derived from GEDI data. This demonstrates MARSNet's enhanced sensitivity in capturing the variability in forest structure across the height spectrum. It is more adept at characterizing forests of very short or tall stature compared to the random forest model.

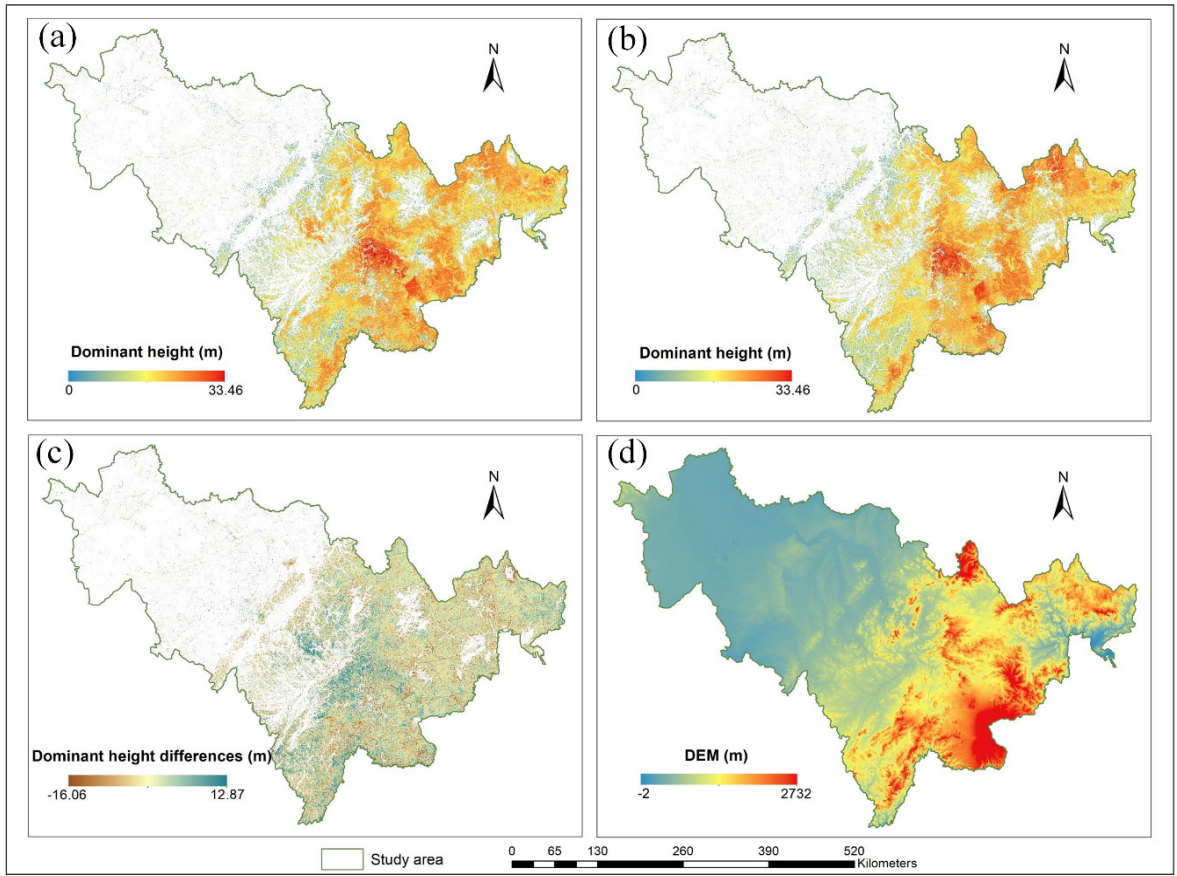

**Fig. 11.** Forest dominant height map of Jilin province derived from different models, difference map between models and DEM. (a) MARSNet, (b) Random forest, (c) Differences between forest dominant height maps produced by MARSNet and random forest, (d) DEM.

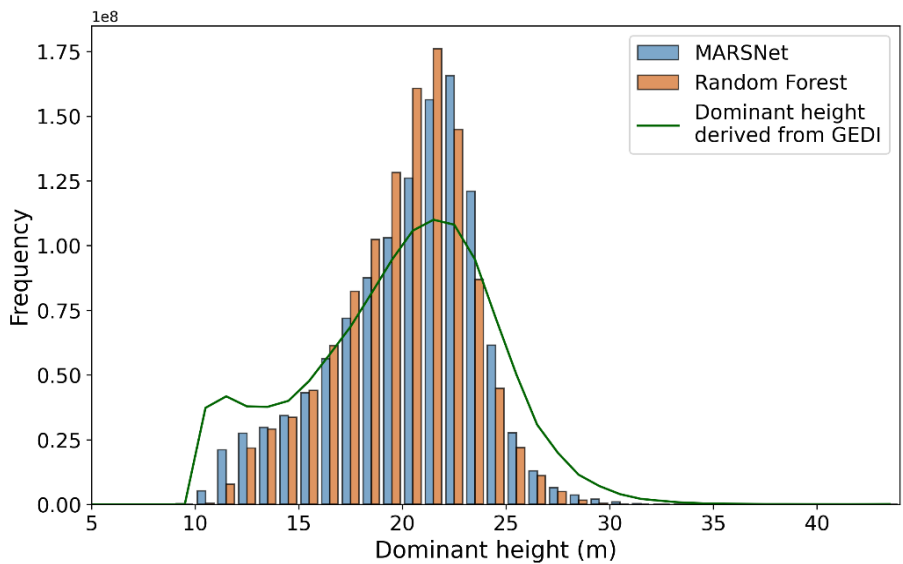

**Fig. 12.** Comparative histograms of dominant heights from MARSNet and random forest with adjusted frequency scaling for dominant heights derived from GEDI.

## 4. Discussion

*4.1. Potential of Multimodal Data Integration and Deep Learning*

In this study, we proposed a novel deep learning model, MARSNet, tailored for forest height estimation from multisource earth observation data. Using GEDI data as labels, MARSNet integrates Sentinel-1 radar, Sentinel-2 multispectral, ALOS-2 PALSAR-2 radar, and ancillary datasets through dedicated encoders, before fusing to predict forest dominant height in a shared decoder. Evaluations on both test and independent field data demonstrate MARSNet's capabilities for accurate dominant height estimation. Through our ablation study of MARSNet modules, we discerned that using distinct encoders for each data type can significantly enhance estimation accuracy. This is attributed to the inherent complexities and unique characteristics within diverse data sources - encoding all modalities together risks undue influence and overfitting from the intricacies of a particular data type. Customized feature extraction overcomes this hurdle by allowing tailored learning of representations aligned with each input. Ablation studies of MARSNet architecture further verified the efficacy of the ESBC modules for forest dominant height estimation. The ESBC components suppresses spatial and band redundancies through reconstruction, while promoting the extraction of informative representations.

While not exhaustively analysing every possible data ablation, the ablation study of remote sensing data modalities provides preliminary evidence of the advantages of utilizing multiple remote sensing data sources. This finding aligns with conclusions drawn in previous research (Ehlers et al. 2022). This is because each data source provides unique and complementary information– optical imagery for spectral signatures, radar backscatter for structural responses, and LiDAR for direct vertical measurements. The heterogenous observations compensate for inherent limitations of individual sensors, allowing more comprehensive characterization of forest structural attributes from multiple perspectives. Owing to the representational superiority of the MARSNet architecture, it achieved slightly higher accuracy over the random forest using all four remote sensing data sources, even when leveraging only Sentinel-2 optical imagery. This underscores the ability of MARSNet for forest height estimation when certain data sources are unavailable, empowering reasonable forest mapping solely from ubiquitous optical satellites when necessary.

Additionally, compared to the conventional machine learning approach of random forest, while MARSNet still exhibits tendencies to overestimate in lower value regions and underestimate in higher value areas, this trend is significantly mitigated. This greatly improves upon the limitations of conventional machine learning methods in effectively modelling at the extreme low and high ends (Pourshamsi et al. 2021; Shendryk 2022). In summary, MARSNet demonstrates strong adaptability, establishing itself as a potent tool for tree height estimation.

*4.2. The relationship between GEDI RH98 and field data*

As observed in Fig. 6, there exists a strong correlation between GEDI RH98 and the field-measured dominant height, enabling us to convert RH98 into dominant height. To further analyse the accuracy of the GEDI data in relation to the field measurements, we inverted the axes as our field data assumes to represent truth (Fig. 13). This reversal facilitates visual assessment and quantification of deviation in the GEDI estimates from the ground-truth field observations. The slope near unity signifies RH98 scales equivalently to true forest height, while the offset of -4.32 indicates RH98 tend to underestimate dominant height. Since the field plots have high sensitivities above 0.97 and canopy cover below 91%, it is feasible for GEDI to detect the ground in these conditions (Hancock et al. 2019). Furthermore, we did not identify any correlation between canopy cover and the deviation of GEDI from field observations. This further substantiates the assertion that the -4.32 m offset is unlikely to be caused by GEDI's ground detection failure. We further split the data based on the ratio of GEDI RH80 to RH98 (Fig. 13). For field plots where the ratio exceeded 0.4, the offset was minimal, at just -0.89. However, for field plots with a ratio less than 0.4, the offset increased substantially to -8.41. This significant offset indicates that the GEDI RH98 considerably underestimated the actual dominant height for these specific points. A smaller ratio of GEDI RH80 to RH98 suggests that energy returned from higher canopy components occupy a smaller fraction of the footprint. Therefore, the large negative offset of 8.41 m for these field plots could stem from sparser higher canopy components, with GEDI's RH98 metric potentially missing the tree tops in these cases. This differentiation based on the ratio underscores the variability in GEDI's accuracy across different forest structures and highlights the importance of considering forest types in the future study.

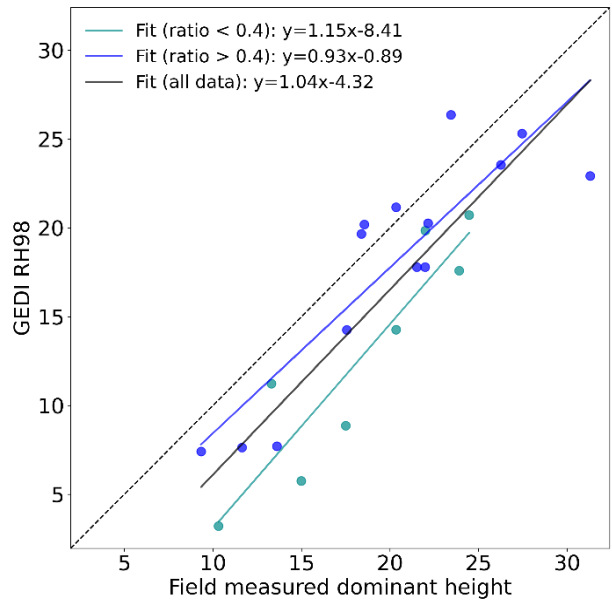

**Fig. 13.** Relationship of field measured dominant height and GEDI RH98. Field plots were divided into two groups based on the ratio of GEDI RH80 to RH98.

Additionally, the geolocation uncertainty of GEDI data can significantly influence its relationship with field measurements. Our field data were collected at the Version 1 footprint locations, having an expected geolocation error of 20.9 m (Beck et al. 2021). The average distance between these 22 field plots and the GEDI Version 2 footprints, which have a smaller geolocation error, is 16.02 m. Although Version 2 still has a geolocation error of 10.3 m (Beck et al. 2021), it suggests that there might be a discrepancy between our field plots and the actual locations of GEDI footprints. Further research could incorporate airborne LiDAR scanning (ALS) data to refine GEDI geolocation and quantify impacts on retrieval accuracy.

It should be noted that there might also be uncertainties associated with the field-measured heights. (Wang et al. 2019). Despite potential uncertainties of both GEDI geolocation and the field measurements, the RH98 metric exhibits a strong relationship with field-measured dominant height. In summary, under present geolocation uncertainties, GEDI demonstrates considerable potential for estimating forest dominant height.

*4.3. Seasonality on forest height estimation*

The forests in Jilin Province consist largely of deciduous trees. Consequently, LiDAR data acquired during leaf-on and leaf-off conditions might exhibit variations due to the changing canopy structure and density across seasons (Hill and Broughton 2009). Some studies have demonstrated that RF models utilizing Airborne LiDAR and multi-seasonal composite imagery yield slightly higher accuracy in predicting canopy height compared to models using median time-series data (Shimizu et al. 2020). However, there is limited literature discussing the impact of using GEDI data from different seasons on forest height estimation. In this study, within the forested area under consideration, although the monthly values of GEDI RH98 exhibited variations, these discrepancies in RH98 were attributed to the differing locations of GEDI footprints for each month, rather than displaying a clear seasonal trend (Fig. 14). Consequently, we did not apply any temporal selection based on the acquisition time of the GEDI data. There is a need for more targeted research to further explore the estimation of forest height by integrating GEDI data from specific seasons with corresponding seasonal remote sensing imagery.

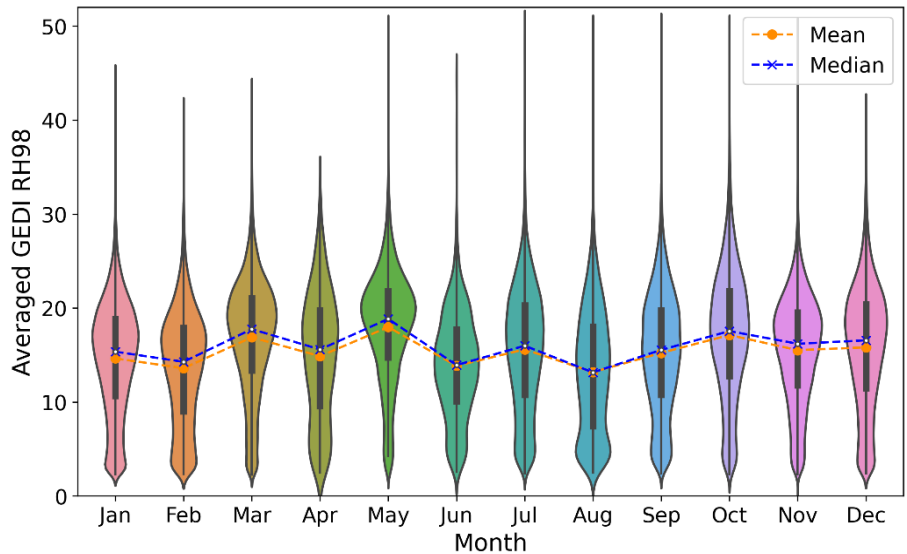

**Fig. 14.** The GEDI RH98 for each month within the forested area of Jilin province in 2021.

## 5. Conclusion

In the study, we proposed a novel deep learning framework, MARSNet, to estimate forest dominant height. Leveraging the availability of field measurements within GEDI footprints, we constructed a robust relationship between the GEDI RH98 and average height of the top 10 trees per plot from field data. This allowed us to derive dominant height from GEDI as labels, and consequently integrate Sentinel-1, Sentinel-2, ALOS-2 PALSAR-2, and ancillary data through dedicated encoders before fusing to predict dominant height in a shared decoder. Comparative experiments against RF, and ablation studies on the data modalities and modules demonstrate the superiority performance of MARSNet. Compared to RF, MARSNet improved the $R^2$ from 0.55 to 0.62 and reduced the RMSE from 3.05 m to 2.82 m on the test data. The ablation experiments also confirmed the pivotal role of multimodality data and the efficacy of separate encoders and ESBConv module for each modality in elevating precision. Even without independent encoders and ESBConv, the deep learning model still outperformed RF in accuracy, underscoring the potential of deep learning for forest height estimation. Applying MARSNet and RF to generate 10 m resolution wall-to-wall maps in Jilin province and validating with field measurements showed MARSNet achieved an $R^2$ of 0.58 and RMSE of 3.76 m, markedly higher than the $R^2$ of 0.41 and RMSE of 4.37 m for RF. The MARSNet demonstrates the immense potential for monitoring forest height. It could facilitate improved quantification of forest structural parameters to support biomass mapping, carbon stocks assessment, inform climate mitigation policies to better cope with climate change.

**Credit author statement**

Conceptualization: W.D., H.Y., E.T.A.M.; Methodology: M.C., W.D., H.Y.; Formal analysis: M.C., W.D., C.M.R., E.T.A.M.; Writing – original draft: M.C., W.D., H.Y.; Software: H.L., S.G.; Data curation: W.D.; Writing – review & editing: I.W., C.M.R., S.G., E.T.A.M..

**Declaration of Competing Interest**

The authors declare that they have no known competing financial interests or personal relationships that could have appeared to influence the work reported in this paper.

**Acknowledgements**

The field work was supported by Davis Expedition Fund, Elizabeth Sinclair Irvine Bequest and Centenary Agroforestry 89 Fund, Moray Endowment Fund and Meiklejohn fund. C.M.R. was supported by the NERC funded

SECO project: NE/T01279X/1. The authors sincerely thank the support from the open access remote sensing datasets that enabled this research, including GEDI data by NASA, Sentinel-1 and Sentinel-2 missions by the European Space Agency (ESA), ALOS-2 PALSAR-2 by the Japan Aerospace Exploration Agency (JAXA), and NASA's High-Resolution Digital Elevation Model (NASADEM).

**Code availability**

The code to train the deep learning models and to predict dominant height will be available at: https://github.com